\documentclass[10pt,twocolumn,letterpaper]{article}

\usepackage{cvpr}
\usepackage{times}
\usepackage{epsfig}
\usepackage{graphicx}
\usepackage{amsmath}
\usepackage{amssymb}
\usepackage{multirow}

\usepackage{bm}
\usepackage[american]{babel}
\usepackage{microtype}
\usepackage{enumitem}

\usepackage{amsfonts}
\usepackage{mathrsfs}
\usepackage{epstopdf}
\usepackage{algorithm}
\usepackage{algorithmic}

\usepackage[pagebackref=true,breaklinks=true,letterpaper=true,colorlinks,bookmarks=false]{hyperref}

\cvprfinalcopy % *** Uncomment this line for the final submission

\def\ie{{\em i.e.}}
\def\eg{{\em e.g.}}
\def\etal{{\em et al.}}

\def\cvprPaperID{5147} % *** Enter the CVPR Paper ID here

\ifcvprfinal\fi

\begin{document}

\title{BBN: Bilateral-Branch Network with Cumulative Learning\\for Long-Tailed Visual Recognition}

\author{Boyan Zhou$^{1}$ \qquad Quan Cui$^{1,2}$ \qquad Xiu-Shen Wei$^{1}$\thanks{Q. Cui and Z.-M. Chen's contribution was made when they were interns in Megvii Research Nanjing, Megvii Technology, China. X.-S. Wei is the corresponding author (weixs.gm@gmail.com).} \qquad Zhao-Min Chen$^{1,3}$\\
\noindent $^1$Megvii Technology \qquad $^2$Waseda University \qquad $^3$Nanjing University\\
%{\tt\small \{zhouboyan94, weixs.gm, chenzhaomin123\}@gmail.com, cui-quan@toki.waseda.jp}\\
}

\maketitle
%\thispagestyle{empty}

%%%%%%%%% ABSTRACT
\begin{abstract}
Our work focuses on tackling the challenging but natural visual recognition task of long-tailed data distribution (\ie, a few classes occupy most of the data, while most classes have rarely few samples). In the literature, class re-balancing strategies (\eg, re-weighting and re-sampling) are the prominent and effective methods proposed to alleviate the extreme imbalance for dealing with long-tailed problems. In this paper, we firstly discover that these re-balancing methods achieving satisfactory recognition accuracy owe to that they could significantly promote the classifier learning of deep networks. However, at the same time, they will unexpectedly damage the representative ability of the learned deep features to some extent. Therefore, we propose a unified Bilateral-Branch Network (BBN) to take care of both representation learning and classifier learning simultaneously, where each branch does perform its own duty separately. In particular, our BBN model is further equipped with a novel cumulative learning strategy, which is designed to first learn the universal patterns and then pay attention to the tail data gradually. Extensive experiments on four benchmark datasets, including the large-scale iNaturalist ones, justify that the proposed BBN can significantly outperform state-of-the-art methods. Furthermore, validation experiments can demonstrate both our preliminary discovery and effectiveness of tailored designs in BBN for long-tailed problems. Our method won the first place in the iNaturalist 2019 large scale species classification competition, and our code is open-source and available at \url{https://github.com/Megvii-Nanjing/BBN}.
\end{abstract}

%%%%%%%%% BODY TEXT

\section{Introduction}

\begin{figure}
\centering
\includegraphics[width=1.0\linewidth]{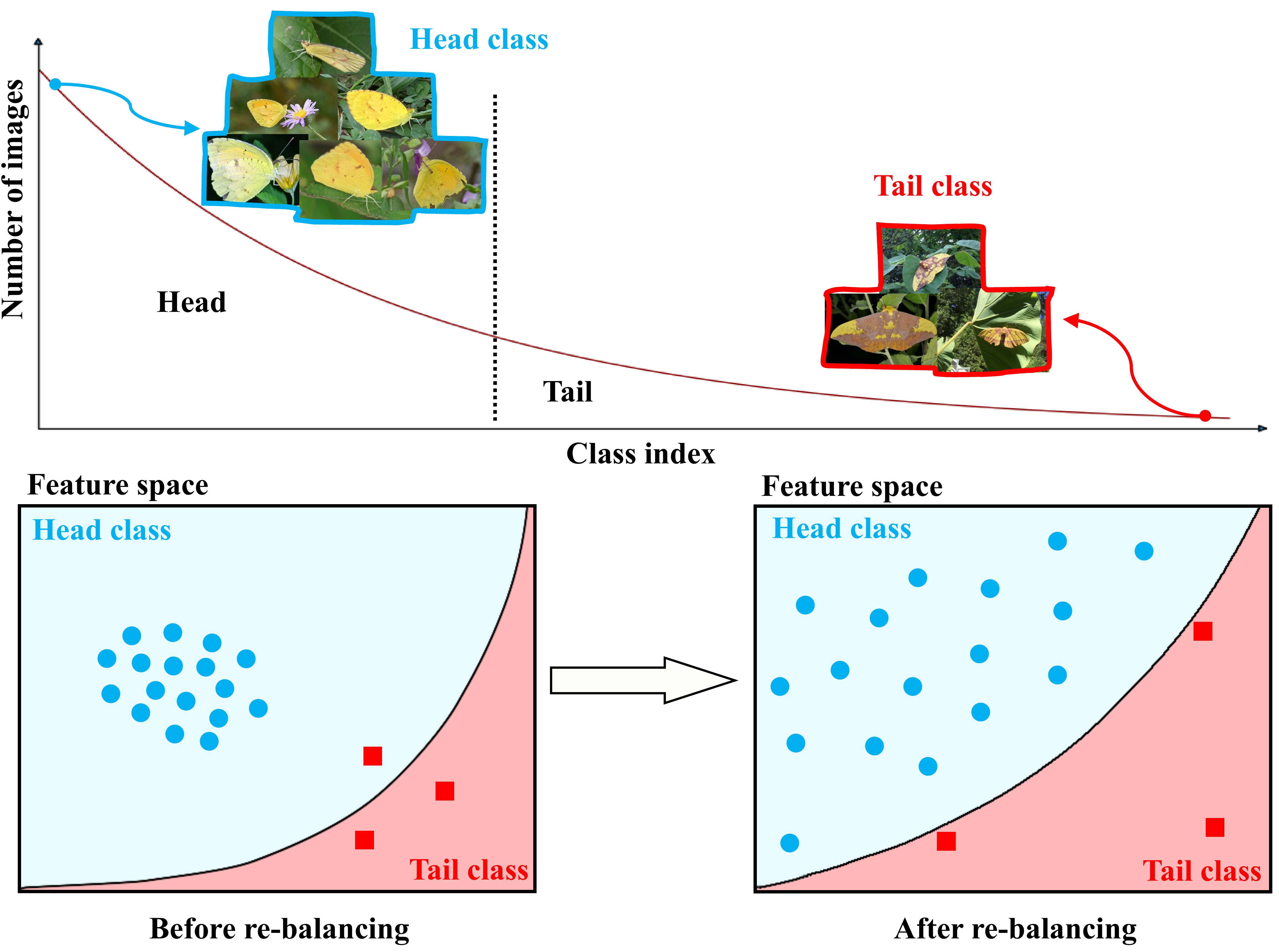}
%\vspace{-2mm}
\caption{Real-world large-scale datasets often display the phenomenon of long-tailed distributions. The extreme imbalance causes tremendous challenges on the classification accuracy, especially for the tail classes. Class re-balancing strategies can yield better classification accuracy for long-tailed problems. In this paper, we reveal that the mechanism of these strategies is to significantly promote classifier learning but will unexpectedly damage the representative ability of the learned deep features to some extent. As conceptually demonstrated, after re-balancing, the decision boundary (\ie, black solid arc) tends to accurately classify the tail data (\ie, red squares). However, the intra-class distribution of each class becomes more separable. Quantitative results are presented in Figure~\ref{fig:representation and classifier learning}, and more analyses can be found in the supplementary materials.}
\label{fig:introduction}
\end{figure}

\begin{figure*}[t]
%\vspace{-9mm}
	\centering
	\begin{minipage}[t]{0.48\textwidth}
		\centering
		\includegraphics[width=6cm]{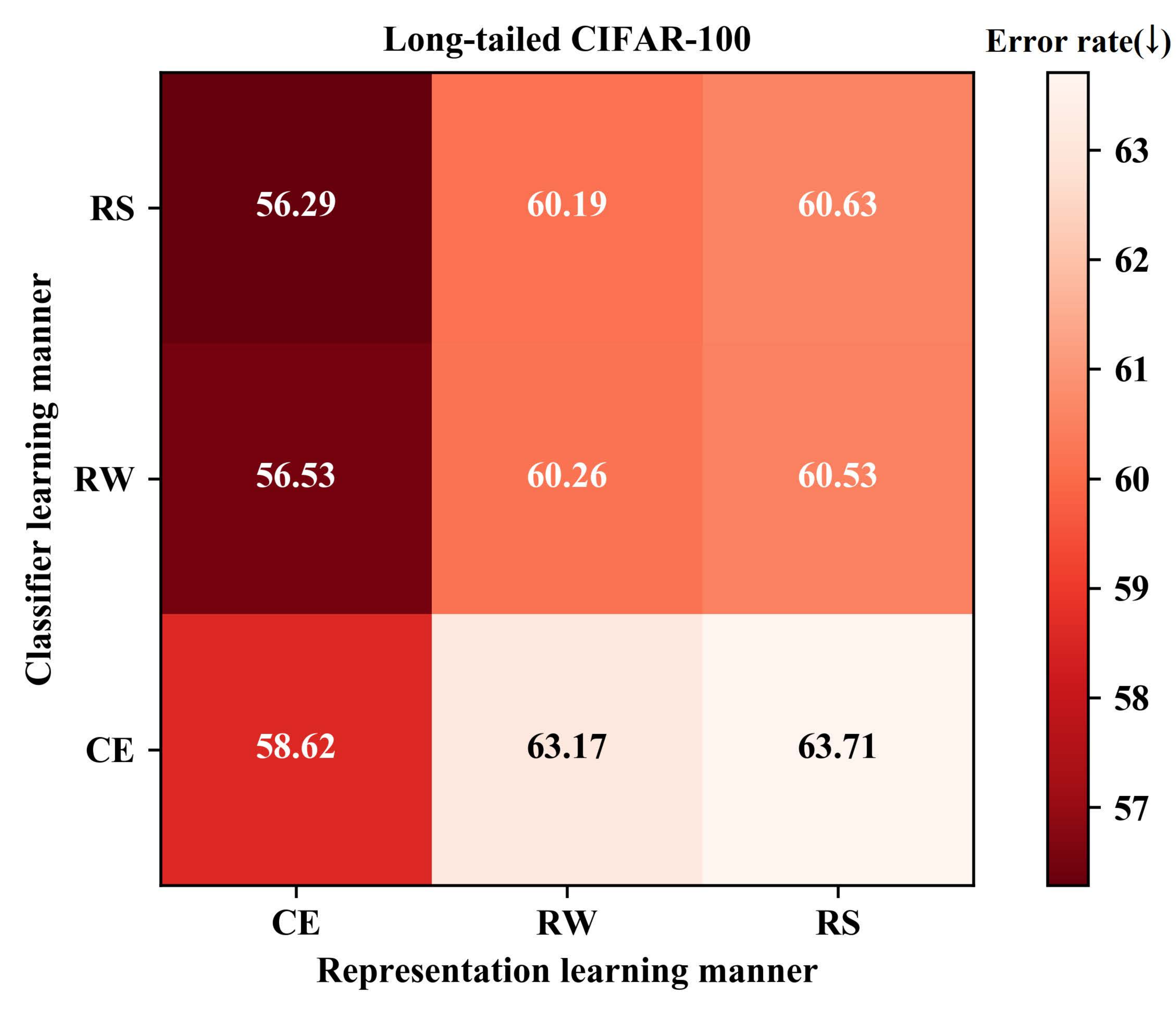}
	\end{minipage}
	\begin{minipage}[t]{0.48\textwidth}
		\centering
		\includegraphics[width=6cm]{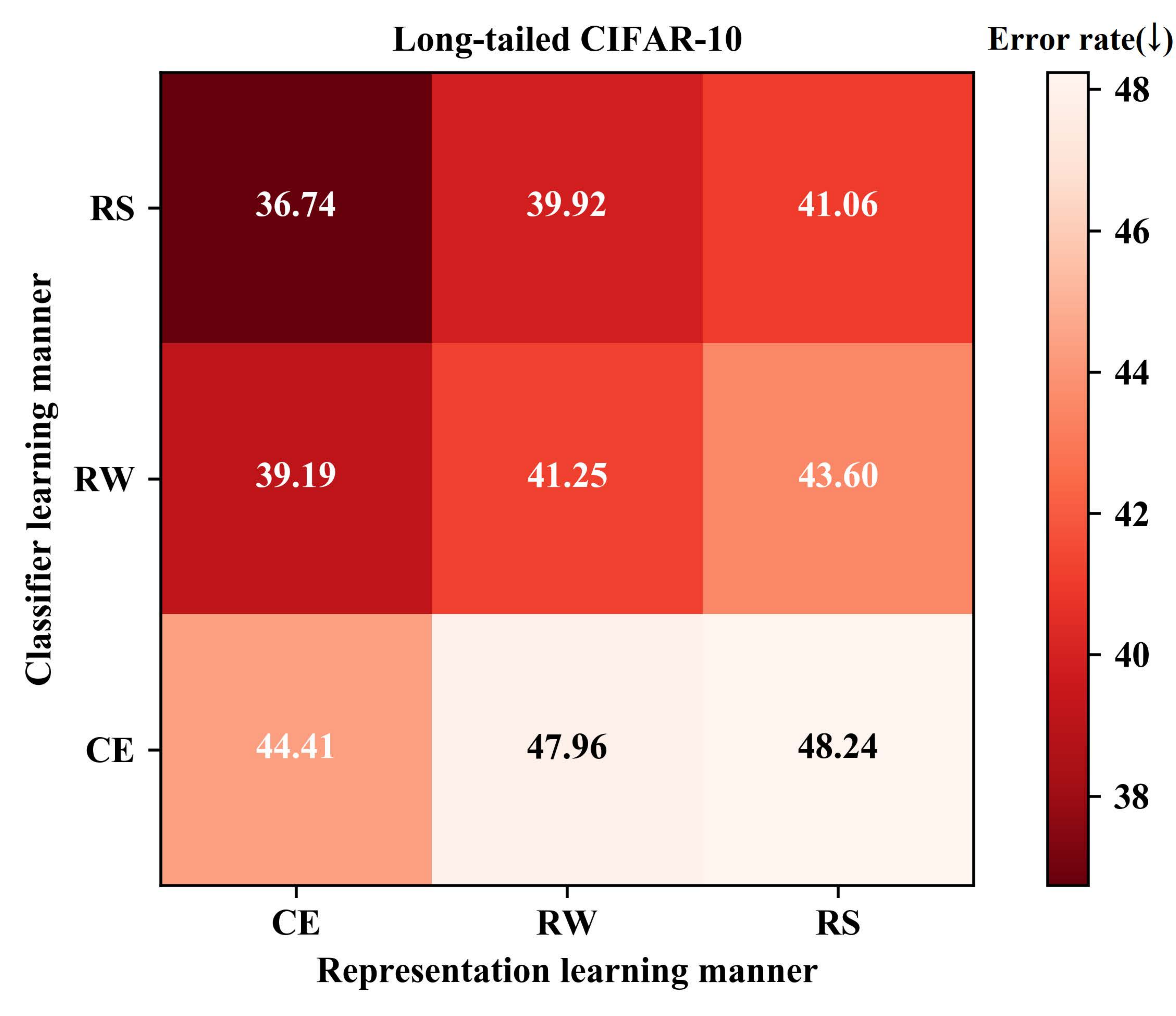}
	\end{minipage}
	\caption{Top-1 error rates of different manners for representation learning and classifier learning on two long-tailed datasets CIFAR-100-IR50 and CIFAR-10-IR50~\cite{ldam}. ``CE'' (Cross-Entropy), ``RW'' (Re-Weighting) and ``RS'' (Re-Sampling) are the conducted learning manners. As observed, when fixing the representation (comparing error rates of three blocks in the vertical direction), the error rates of classifiers trained with RW/RS are reasonably lower than CE. While, when fixing the classifier (comparing error rates in the horizontal direction), the representations trained with CE surprisingly get lower error rates than those with RW/RS. Experimental details can be found in Section~\ref{sec:how class re-balancing work}.}
	\label{fig:representation and classifier learning}
\end{figure*}

With the advent of research on deep Convolutional Neural Networks (CNNs), the performance of image classification has witnessed incredible progress. The success is undoubtedly inseparable to available and high-quality large-scale datasets, \eg, ImageNet ILSVRC 2012~\cite{imagenet}, MS COCO~\cite{coco} and Places Database~\cite{zhou2017places}, \etc. In contrast with these visual recognition datasets exhibiting roughly uniform distributions of class labels, real-world datasets always have skewed distributions with \emph{a long tail}~\cite{kendall1948advanced,van2017devil}, \ie, a few classes (a.k.a. \emph{head class}) occupy most of the data, while most classes (a.k.a. \emph{tail class}) have rarely few samples, cf.~Figure~\ref{fig:introduction}. Moreover, more and more long-tailed datasets reflecting the realistic challenges are constructed and released by the computer vision community in very recent years, \eg, iNaturalist~\cite{cui2018large}, LVIS~\cite{Gupta2019LVIS} and RPC~\cite{wei2019rpc}. When dealing with such visual data, deep learning methods are not feasible to achieve outstanding recognition accuracy due to both the data-hungry limitation of deep models and also the extreme class imbalance trouble of long-tailed data distributions.

In the literature, the prominent and effective methods for handling long-tailed problems are class re-balancing strategies, which are proposed to alleviate the extreme imbalance of the training data. Generally, class re-balancing methods are roughly categorized into two groups, \ie, re-sampling~\cite{shen2016relay,buda2018systematic,japkowicz2002class,buda2018systematic,he2009learning,byrd2019effect,drummond2003c4,more2016survey,chawla2002smote} and cost-sensitive re-weighting~\cite{huang2016learning,wang2017learning,cb-focal,ren18l2rw}. These methods can adjust the network training, by re-sampling the examples or re-weighting the losses of examples within mini-batches, which is in expectation closer to the test distributions. Thus, class re-balancing is effective to directly influence the classifier weights' updating of deep networks, \ie, promoting the classifier learning. That is the reason why re-balancing could achieve satisfactory recognition accuracy on long-tailed data.

\begin{figure*}[t]
	\centering
	%\vspace{-0mm}
	\includegraphics[width=0.7\linewidth]{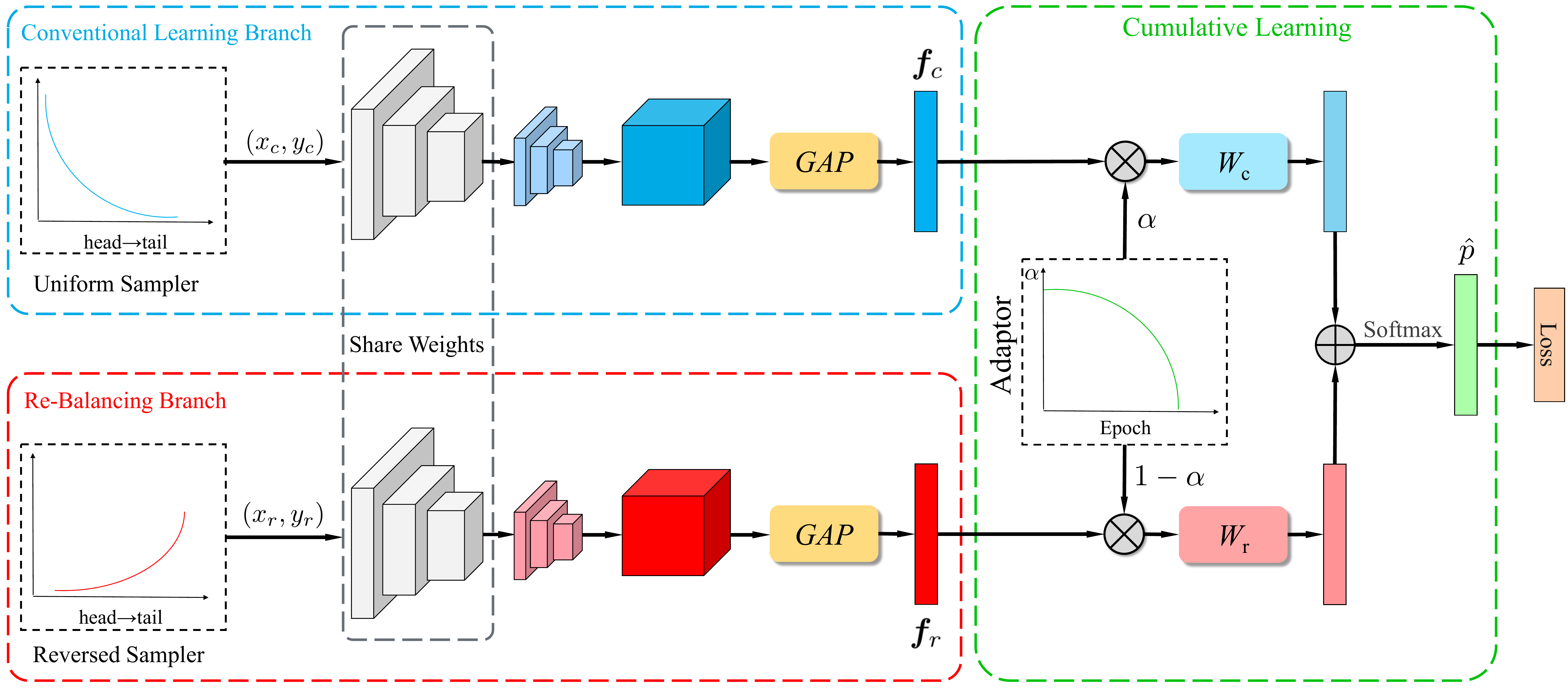}
	%\vspace{-2mm}
	\caption{Framework of our Bilateral-Branch Network (BBN). It consists of three key components: 1) The \emph{conventional learning branch} takes input data from a uniform sampler, which is responsible for learning universal patterns of original distributions. While, 2) the \emph{re-balancing branch} takes inputs from a reversed sampler and is designed for modeling the tail data. The output feature vectors $\bm{f}_c$ and $\bm{f}_r$ of two branches are aggregated by 3) our \emph{cumulative learning strategy} for computing training losses. ``\emph{GAP}'' is short for global average pooling.}
	\label{fig:structure}%Both branches are the structure of residual network~\cite{he2016deep}, \eg, ResNet-50, and share the same parameters except the last residual block.
\end{figure*}

However, although re-balancing methods have good eventual predictions, we argue that these methods still have adverse effects, \ie, they will also unexpectedly damage the representative ability of the learned deep features (\ie, the representation learning) to some extent. In concretely, re-sampling has the risks of over-fitting the tail data (by over-sampling) and also the risk of under-fitting the whole data distribution (by under-sampling), when data imbalance is extreme. For re-weighting, it will distort the original distributions by directly changing or even inverting the data presenting frequency.

As a preliminary of our work, by conducting validation experiments, we justify our aforementioned argumentations. Specifically, to figure out how re-balancing strategies work, we divide the training process of deep networks into two stages, \ie, to separately conduct the representation learning and the classifier learning. At the former stage for representation learning, we employ plain training (conventional cross-entropy), re-weighting and re-sampling as three learning manners to obtain their corresponding learned representations. Then, at the latter stage for classifier learning, we first fix the parameters of representation learning (\ie, backbone layers) converged at the former stage and then retrain the classifiers of these networks (\ie, fully-connected layers) \emph{from scratch}, also with the three aforementioned learning manners. In Figure~\ref{fig:representation and classifier learning}, the prediction error rates on two benchmark long-tailed datasets~\cite{ldam}, \ie, CIFAR-100-IR50 and CIFAR-10-IR50, are reported. Obviously, when fixing the representation learning manner, re-balancing methods reasonably achieve lower error rates, indicating they can promote classifier learning. On the other side, by fixing the classifier learning manner, plain training on original imbalanced data can bring better results according to its better features. Also, the worse results of re-balancing methods prove that they will hurt feature learning.

Therefore, in this paper, for exhaustively improving the recognition performance of long-tailed problems, we propose a unified Bilateral-Branch Network (BBN) model to take care of \emph{both representation learning and classifier learning} simultaneously. As shown in Figure~\ref{fig:structure}, our BBN model consists of two branches, termed as the ``conventional learning branch'' and the ``re-balancing branch''. In general, each branch of BBN separately performs its own duty for representation learning and classifier learning, respectively. As the name suggests, the conventional learning branch equipped with the typical uniform sampler w.r.t. the original data distribution is responsible for learning universal patterns for recognition. While, the re-balancing branch coupled with a reversed sampler is designed to model the tail data. After that, the predicted outputs of these bilateral branches are aggregated in the cumulative learning part by an adaptive trade-off parameter $\alpha$. $\alpha$ is automatically generated by the ``{Adaptor}'' according to the number of training epochs, which adjusts the whole BBN model to firstly learn the universal features from the original distribution and then pay attention to the tail data gradually. More importantly, $\alpha$ could further control the parameter updating of each branch, which, for example, avoids damaging the learned universal features when emphasizing the tail data at the later periods of training.

In experiments, empirical results on four benchmark long-tailed datasets show that our model obviously outperforms existing state-of-the-art methods. Moreover, extensive validation experiments and ablation studies can prove the aforementioned preliminary discovery and also validate the effectiveness of our tailored designs for long-tailed problems.

The main contributions of this paper are as follows:
\begin{itemize}[itemsep=-0.2em, leftmargin=1em]
	\item We explore the mechanism of the prominent class re-balancing methods for long-tailed problems, and further discover that these methods can significantly promote classifier learning and meanwhile will affect the representation learning w.r.t. the original data distribution.
	\item We propose a unified Bilateral-Branch Network (BBN) model to take care of both representation learning and classifier learning for exhaustively boosting long-tailed recognition. Also, a novel cumulative learning strategy is developed for adjusting the bilateral learnings and coupled with our BBN model's training.
	\item We evaluate our model on four benchmark long-tailed visual recognition datasets, and our proposed model consistently achieves superior performance over previous competing approaches.
\end{itemize}

\section{Related work}

\textbf{Class re-balancing strategies:} \emph{Re-sampling} methods as one of the most important class re-balancing strategies could be divided into two types: 1) Over-sampling by simply repeating data for minority classes~\cite{shen2016relay, buda2018systematic,  byrd2019effect} and 2) under-sampling by abandoning data for dominant classes~\cite{japkowicz2002class,buda2018systematic,he2009learning}. But sometimes, with re-sampling, duplicated tailed samples might lead to over-fitting upon minority classes~\cite{chawla2002smote,cb-focal}, while discarding precious data will certainly impair the generalization ability of deep networks.

\emph{Re-weighting} methods are another series of prominent class re-balancing strategies, which usually allocate large weights for training samples of tail classes in loss functions~\cite{huang2016learning, wang2017learning}. However, re-weighting is not capable of handling the large-scale, real-world scenarios of long-tailed data and tends to cause optimization difficulty~\cite{mikolov2013distributed}. Consequently, Cui~\etal~\cite{cb-focal} proposed to adopt the effective number of samples~\cite{cb-focal} instead of proportional frequency. Thereafter, Cao~\etal~\cite{ldam} explored the margins of the training examples and designed a label-distribution-aware loss to encourage larger margins for minority classes.

%Nevertheless, although these two kinds of class-rebalancing methods could benefit the final prediction when facing long-tailed data, both of them bring bad influence in feature learning as aforementioned. In this paper, we conduct contrast experiments on long-tailed CIFAR~\cite{ldam} to figure out how these re-balancing strategies work to boost the final performance, cf. Section~\ref{sec:how class re-balancing work}.

In addition, recently, some two-stage fine-tuning strategies~\cite{ldam, cui2018large, ouyang2016factors} were developed to modify re-balancing for effectively handling long-tailed problems. Specifically, they separated the training process into two single stages. In the first stage, they trained networks as usual on the original imbalanced data and only utilized re-balancing at the second stage to fine-tune the network with a small learning rate.

Beyond that, other methods of different learning paradigms were also proposed to deal with long-tailed problems, \eg, metric learning~\cite{zhang2017range, huang2016learning}, meta-learning~\cite{Liu_2019_CVPR} and knowledge transfer learning~\cite{wang2017learning,Zhong_2019_CVPR}, which, however, are not within the scope of this paper.

\textbf{Mixup:}
Mixup~\cite{mixup} was a general data augmentation algorithm, \ie, convexly combining random pairs of training images and their associated labels, to generate \emph{additional} samples when training deep networks. Also, manifold mixup~\cite{manifoldmixup} conducted mixup operations on random pairs of samples in the manifold feature space for augmentation. The mixed ratios in mixup were sampled from the $\beta$-distribution to increase the randomness of augmentation. Although mixup is clearly far from our unified end-to-end trainable model, in experiments, we still compared with a series of mixup algorithms to validate our effectiveness.

\section{How class re-balancing strategies work?}
\label{sec:how class re-balancing work}

%There is a heuristic observation proposed in the previous work~\cite{ldam} that deep feature representations trained with re-sampling/re-weighting strategy are not as expressive as those trained with vanilla classification loss.  This observation inspires us to figure out a question: since the re-sampling/re-weighting strategy could damage the representation, what is the way these strategies work to promote the final classification performance? It motivates us with a good starting point where we should explore what exactly re-sampling and re-weighting influence.

%As aforementioned, re-balancing strategies could still have adverse effects. 
%Specifically, re-sampling has the risks of over-fitting the tail data (by over-sampling) and also the risk of under-fitting the whole data distribution (by under-sampling), when data imbalance is extreme. For re-weighting, it will distort the original distributions by directly changing or even inverting the data presenting frequency. 
%Since bad influence could be introduced by these strategies, in this section, we explore how these strategies work to boost the final classification performance on long-tailed datasets.

In this section, we attempt to figure out the working mechanism of these class re-balancing methods. More concretely, we divide a deep classification model into two essential parts: 1) the feature extractor (\ie, frontal base/backbone networks) and 2) the classifier (\ie, last fully-connected layers). Accordingly, the learning process of a deep classification network could be separated into representation learning and classifier learning. Since class re-balancing strategies could boost the classification accuracy by altering the training data distribution closer to test and paying more attention to the tail classes, we propose a conjecture that the way these strategies work is to promote classifier learning significantly but might damage the universal representative ability of the learned deep features due to distorting original distributions.

%\ie, in the representation learning stage, the conventional training strategy can exert better performance, while in the classifier learning stage re-sampling/re-weighting strategy tends to be better.

%As illustrated in Figure~\ref{fig:introduction}, the model trained on the original long-tailed distribution could suffer from extreme class imbalance (head class oridinary car \emph{v.s} tail class luxury car) and get poor performance on tail classes. Existing re-weighting and re-sampling strategies could improve the performance of classifiers but decrease the quality of features. 
%Hence, in this section, we attempt to explore how re-sampling and re-weighting strategies influence representation learning and classifier learning respectively. We learn representations and classifiers in a two-stage manner. 

In order to justify our conjecture, we design a two-stage experimental fashion to separately learn representations and classifiers of deep models. Concretely, in the first stage, we train a classification network with plain training (\ie, cross-entropy) or re-balancing methods (\ie, re-weighting/re-sampling) as learning manners. Then, we obtain different kinds of feature extractors corresponding to these learning manners. When it comes to the second stage, we fix the parameters of the feature extractors learned in the former stage, and retrain classifiers \textit{from scratch} with the aforementioned learning manners again. In principle, we design these experiments to fairly compare the quality of representations and classifiers learned by different manners by following the control variates method.

%If representations are invariable and the classification performance becomes better when we switch the learning manner of classifiers from cross-entropy to re-weighting/re-sampling, our conjecture concerning classifier learning is proven to be correct. For the representation learning conjecture, based on the assumption that both classifiers are trained with the same strategy, if classification performance on representations trained with cross-entropy is better, our conjecture on representation learning will be supported.

The CIFAR~\cite{cifar} datasets are a collection of images that are commonly used to assess computer vision approaches. Previous work~\cite{cb-focal,ldam} created long-tailed versions of CIFAR datasets with different imbalance ratios, \ie, the number of the most frequent class divided by the least frequent class, to evaluate the performance. In this section, following~\cite{ldam}, we also use long-tailed CIFAR-10/CIFAR-100 as the test beds.

As shown in Figure~\ref{fig:representation and classifier learning}, we conduct several contrast experiments to validate our conjecture on CIFAR-100-IR50 (long-tailed CIFAR-100 with imbalance ratio $50$). As aforementioned, we separate the whole network into two parts: the feature extractor and classifier. Then, we apply three manners for the feature learning and the classifier learning respectively according to our two-stage training fashion. Thus, we can obtain nine groups of results based on different permutations: (1) Cross-Entropy (CE): We train the networks as usual on the original imbalanced data with the conventional cross-entropy loss. (2) Re-Sampling (RS): We first sample a class uniformly and then collect an example from that class by sampling with replacement. By repeating this process, a balanced mini-batch data is obtained. (3) Re-Weighting (RW): We re-weight all the samples by the inverse of the sample size of their classes. The error rate is evaluated on the validation set. As shown in Figure~\ref{fig:representation and classifier learning}, we have the observations from two perspectives:
\begin{itemize}[itemsep=-0.2em, leftmargin=1em]
        \item \textbf{Classifiers:} When we apply the same representation learning manner (comparing error rates of three blocks in the vertical direction), it can be reasonably found that RW/RS always achieve lower classification error rates than CE, which owes to their re-balancing operations adjusting the classifier weights' updating to match test distributions.
	\item \textbf{Representations:} When applying the same classifier learning manner (comparing error rates of three blocks in the horizontal direction), it is a bit of surprise to see that error rates of CE blocks are consistently lower than error rates of RW/RS blocks. The findings indicate that training with CE achieves better classification results since it obtains better features. The worse results of RW/RS reveal that they lead to inferior discriminative ability of the learned deep features.
\end{itemize}

Furthermore, as shown in Figure~\ref{fig:representation and classifier learning} (left), by employing CE on the representation learning and employing RS on the classifier learning, we can achieve the lowest error rate on the validation set of CIFAR-100-IR50. Additionally, to evaluate the generalization ability for representations produced by three manners, we utilize pre-trained models trained on CIFAR-\emph{100}-IR50 as the feature extractor to obtain the representations of CIFAR-\emph{10}-IR50, and then perform the classifier learning experiments as the same as aforementioned. As shown in Figure~\ref{fig:representation and classifier learning} (right), on CIFAR-10-IR50, it can have the identical observations, even in the situation that the feature extractor is trained on another long-tailed dataset.

%------------------------------------------------------------------------
\section{Methodology}

%In this section, we firstly demonstrate the framework of our proposed Bilateral-Branch Network (BBN) and introduce notations used in this paper. Then, we elaborate our method with its main aspects, \ie, (1) bilateral-branch architecture and (2) cumulative learning strategy, in following subsections. 
%Since re-balancing strategies could promote classifier learning but damage representation learning, we propose a simple but effective method to subtly integrate the conventional training strategy and the re-balancing strategy together, called \emph{Bilateral-Branch Network (BBN)}.

\subsection{Overall framework}

As shown in Figure~\ref{fig:structure}, our BBN consists of three main components. Concretely, we design two branches for representation learning and classifier learning, termed ``\textit{conventional learning branch}'' and ``\textit{re-balancing branch}'', respectively. Both branches use the same residual network structure~\cite{he2016deep} and share all the weights except for the last residual block. Let $\mathbf{x}_{\cdot}$ denote a training sample and ${y}_{\cdot}\in\{1,2,\ldots,C\}$ is its corresponding label, where $C$ is the number of classes. For the bilateral branches, we apply uniform and reversed samplers to each of them separately and obtain two samples $(\mathbf{x}_{c}, {y}_{c})$ and $(\mathbf{x}_{r}, {y}_{r})$ as the input data, where $(\mathbf{x}_{c}, {y}_{c})$ is for the conventional learning branch and $(\mathbf{x}_{r}, {y}_{r})$ is for the re-balancing branch. Then, two samples are fed into their own corresponding branch to acquire the feature vectors $\bm{f}_c \in \mathbb{R}^{D}$ and $\bm{f}_r \in \mathbb{R}^{D}$ by global average pooling.

Furthermore, we also design a specific cumulative learning strategy for shifting the learning ``attention'' between two branches in the training phase. In concretely, by controlling the weights for $\bm{f}_c$ and $\bm{f}_r$ with an adaptive trade-off parameter $\alpha$, the weighted feature vectors $\alpha \bm{f}_c$ and $(1-\alpha) \bm{f}_r$ will be sent into the classifiers $\bm{W}_c \in \mathbb{R}^{D\times C}$ and $\bm{W}_r \in \mathbb{R}^{D\times C}$ respectively and the outputs will be integrated together by element-wise addition. The output logits are formulated as
\begin{equation}
\label{output}
\mathbf{z} = \alpha \bm{W}^{\top}_c \bm{f}_c + (1-\alpha) \bm{W}^{\top}_r  	\bm{f}_r,
\end{equation}
where $\mathbf{z}  \in \mathbb{R}^{C}$ is the predicted output, \ie, $[z_1,z_2,\ldots, z_C ]^{\top}$. For each class $i \in \left\lbrace 1, 2, \ldots, C\right\rbrace$, the softmax function calculates the probability of the class by
\begin{equation}
\label{softmax}
\hat{p}_i = \dfrac{e^{z_i}}{\sum_{j=1}^{C} e^{z_j}}.
\end{equation}

Then, we denote $E(\cdot,\cdot)$ as the cross-entropy loss function and the output probability distribution as $\hat{\bm{p}}=[\hat{p}_1,\hat{p}_2,..., \hat{p}_C ]^{\top}$. Thus, the weighted cross-entropy classification loss of our BBN model is illustrated as
\begin{equation}
\label{loss}
\mathcal{L} = \alpha  E({\hat{\bm{p}}, {y}_{c}}) + (1-\alpha)  E({\hat{\bm{p}}, {y}_{r}}), 
\end{equation}
and the whole network is end-to-end trainable.
% Refer to Algorithm~\ref{alg:bbn} for details of training loss computation. 

\subsection{Proposed bilateral-branch structure}

%Re-weighting and re-sampling are the most well-known techniques for dealing with long-tail datasets whose training data distribution is imbalanced while test sets is balanced.
%These techniques try to keep the distributions of training set and test set consistent by applying large weights to tail training samples or sampling tail training samples frequently. 

%As aforementioned in Section~\ref{sec:how class re-balancing work}, directly balancing the training data distribution probably brings bad influences into the representation learning process. To alleviate this problem, Cui \etal~\cite{cui2018large} and Cao \etal~\cite{ldam} proposed to train the network with a two-stage training strategy, \ie, they train the network as usual with the original long-tailed dataset at the first stage, then re-sampling or re-weighting is utilized at the second stage. Noting that the second stage still tries to balance the training data distribution, resulting in that the representations learned in the first stage are destroyed.

In this section, we elaborate the details of our unified bilateral-branch structure shown in Figure~\ref{fig:structure}. As aforementioned, the proposed {conventional learning branch} and {re-balancing branch} do perform their own duty (\ie, representation learning and classifier learning, respectively). There are two unique designs for these branches.
%As shown in Figure~\ref{fig:structure}, we term these two branches as \textit{conventional learning branch} and \textit{re-balancing branch}. 
%Specifically, two branches are designed for the representation learning and distribution balancing, respectively. Considering that re-balancing will cause damage to representation learning, we attempt to conduct these two tasks simultaneously by proposed dual-branch strategy.

{\textbf{Data samplers. }} 
The input data for the conventional learning branch comes from a uniform sampler, where each sample in the training dataset is sampled only once with equal probability in a training epoch. The uniform sampler retains the characteristics of original distributions, and therefore benefits the representation learning. While, the re-balancing branch aims to alleviate the extreme imbalance and particularly improve the classification accuracy on tail classes~\cite{van2017devil}, whose input data comes from a reversed sampler. For the reversed sampler, the sampling possibility of each class is proportional to the reciprocal of its sample size, \ie, the more samples in a class, the smaller sampling possibility that class has. In formulations, let denote that the number of samples for class $i$ is $N_i$ and the maximum sample number of all the classes is $N_{max}$. There are three sub-procedures to construct the reversed sampler: 1) Calculate the sampling possibility $P_i$ for class $i$ according to the number of samples as
\begin{equation}
\label{weight}
\begin{aligned}
P_i = \frac{w_i}{\sum_{j=1}^{C} w_j}\,,
\end{aligned}
\end{equation}
where $w_i=\frac{N_{max}}{N_i}$; 2) Randomly sample a class according to $P_i$; 3) Uniformly pick up a sample from class $i$ with replacement. By repeating this reversed sampling process, training data of a mini-batch is obtained.

{\textbf{Weights sharing. }} 
In BBN, both branches economically share the same residual network structure, as illustrated in Figure~\ref{fig:structure}. We use ResNets~\cite{he2016deep} as our backbone network, \eg, ResNet-32 and ResNet-50. In details, two branch networks, except for the last residual block, share the same weights. There are two benefits for sharing weights: On the one hand, the well-learned representation by the conventional learning branch can benefit the learning of the re-balancing branch. On the other hand, sharing weights will largely reduce computational complexity in the inference phase.

\subsection{Proposed cumulative learning strategy}\label{sec:cumulative}

Cumulative learning strategy is proposed to shift the learning focus between the bilateral branches by controlling both the weights for features produced by two branches and the classification loss $\mathcal{L}$. It is designed to first learn the universal patterns and then pay attention to the tail data gradually. In the training phase, the feature $\bm{f}_c$ of the conventional learning branch will be multiplied by $\alpha$ and the feature $\bm{f}_r$ of the re-balancing branch will be multiplied by $1-\alpha$, where $\alpha$ is automatically generated according to the training epoch. Concretely, the number of total training epochs is denoted as $T_{max}$ and the current epoch is $T$. $\alpha$ is calculated by
\begin{equation}
\label{alpha}
\alpha = 1 - \left(\frac{T}{T_{max}}\right)^{2},
\end{equation}
which $\alpha$ will gradually decrease as the training epochs increasing.

%Then, the weighted feature representations, $\alpha \bm{f}_c$ and $(1-\alpha)\bm{f}_r$, will be sent into the classifier $\bm{W}_c$ and $\bm{W}_r$ of these bilateral branches, respectively, to produce the prediction logits. Then, the output logits of two classifiers will be integrated by an element-wise addition. 

\begin{table*}[t]
\footnotesize
%\vspace{-4mm}
\caption{Top-1 error rates of ResNet-32 on long-tailed CIFAR-10 and CIFAR-100. (Best results are marked in bold.) }
\vspace{2mm}
\begin{center}
	\begin{tabular}{|c|c|c|c|c|c|c|}
		\hline
		Dataset                      & \multicolumn{3}{c|}{Long-tailed CIFAR-10}                             & \multicolumn{3}{c|}{Long-tailed CIFAR-100}                             \\ \hline
		Imbalance ratio              & \multicolumn{1}{c|}{100} & \multicolumn{1}{c|}{50} & 10             & \multicolumn{1}{c|}{100} & \multicolumn{1}{c|}{50} & 10      \\ 
		\hline
		\hline
		CE                           & 29.64                    & 25.19                   & 13.61          & 61.68                    & 56.15                   & 44.29          \\
		Focal~\cite{focalloss}                        & 29.62                    & 23.28                   & 13.34          & 61.59                    & 55.68                   & 44.22          \\
		Mixup~\cite{mixup}                     & 26.94                    & 22.18                   & 12.90          & 60.46                    & 55.01                   & 41.98          \\
		Manifold Mixup~\cite{manifoldmixup}              & 27.04                    & 22.05                   & 12.97          & 61.75                    & 56.91                   & 43.45          \\
		Manifold Mixup (two samplers) & 26.90                    & 20.79                   & 13.17          & 63.19                    & 57.95                   & 43.54          \\ \hline
		CE-DRW~\cite{ldam}                    & 23.66                    & 20.03                   & 12.44           & 58.49                    & 54.71                   & 41.88         \\
		CE-DRS~\cite{ldam}                     & 24.39                    & 20.19                   & 12.62          & 58.39                    & 54.52                  & 41.89          \\ \hline
		CB-Focal~\cite{cb-focal}                     & 25.43                    & 20.73                   & 12.90           & 60.40                     & 54.83                   & 42.01          \\
		LDAM-DRW~\cite{ldam}                     & 22.97                    & 18.97                   & 11.84          & 57.96                    & 53.38                   & 41.29          \\ \hline
		\hline
		Our BBN                        & \textbf{20.18}           & \textbf{17.82}          & \textbf{11.68} & \textbf{57.44}           & \textbf{52.98}          & \textbf{40.88} \\ \hline
	\end{tabular}
\end{center}
\label{tab:cifar_results}
\end{table*}

In intuition, we design the adapting strategy for $\alpha$ based on the motivation that discriminative feature representations are the foundation for learning robust classifiers. Although representation learning and classifier learning deserve equal attentions, the learning focus of our BBN should gradually change from feature representations to classifiers, which can exhaustively improve long-tailed recognition accuracy. With $\alpha$ decreasing, the main emphasis of BBN turns from the conventional learning branch to the re-balancing branch. Different from two-stage fine-tuning strategies~\cite{ldam, cui2018large, ouyang2016factors}, our $\alpha$ ensures that both branches for different goals can be constantly updated in the whole training process, which could avoid the affects on one goal when it performs training for the other goal.

In experiments, we also provide the qualitative results of this intuition by comparing different kinds of adaptors, cf. Section~\ref{sec:alpha}.	

\subsection{Inference phase}
During inference, the test samples are fed into both branches and two features $\bm{f}'_c$ and $\bm{f}'_r$ are obtained. Because both branches are equally important, we simply fix $\alpha$ to $0.5$ in the test phase. Then, the equally weighted features are fed to their corresponding classifiers (\ie, $\bm{W}_c$ and $\bm{W}_r$) to obtain two prediction logits. Finally, both logits are aggregated by element-wise addition to return the classification results.%Additionally, because only the last residual block is not shared for two branches in our BBN, the computational complexity just increases slightly in the inference phase, while the classification accuracy improves by a large margin.

%------------------------------------------------------------------------
\section{Experiments}
%We evaluate our proposed BBN network on artificially created long-tailed CIFAR~\cite{cifar} datasets based on different imbalance ratios, real-world long-tailed datasets iNaturalist 2017 and iNaturalist 2018~\cite{cui2018large}. Comprehensive ablation studies on our proposed BBN are presented, as well as visualization results.

\subsection{Datasets and empirical settings}\label{sec:datasets}

{\noindent \textbf{Long-tailed CIFAR-10 and CIFAR-100. }} 
Both CIFAR-10 and CIFAR-100 contain 60,000 images, 50,000 for training and 10,000 for validation with category number of 10 and 100, respectively. For fair comparisons, we use the long-tailed versions of CIFAR datasets as the same as those used in~\cite{ldam} with controllable degrees of data imbalance. We use an imbalance factor $\beta$ to describe the severity of the long tail problem with the number of training samples for the most frequent class and the least frequent class, \eg, $\beta=\frac{N_{max}}{N_{min}}$. Imbalance factors we use in experiments are $10$, $50$ and $100$.

{\noindent \textbf{iNaturalist 2017 and iNaturalist 2018. }} 
The iNaturalist species classification datasets are large-scale real-world datasets that suffer from extremely imbalanced label distributions. The 2017 version of iNaturalist contains 579,184 images with 5,089 categories and the 2018 version is composed of 437,513 images from 8,142 categories. Note that, besides the extreme imbalance, the iNaturalist datasets also face the fine-grained problem~\cite{wei2019deep,Zhao2017,scda,pcm}. In this paper, the official splits of training and validation images are utilized for fair comparisons.

\subsection{Implementation details}

{\noindent \textbf{Implementation details on CIFAR. }}  
For long-tailed CIFAR-10 and CIFAR-100 datasets, we follow the data augmentation strategies proposed in~\cite{he2016deep}: randomly crop a $32\times32$ patch from the original image or its horizontal flip with $4$ pixels padded on each side. We train the ResNet-32~\cite{he2016deep} as our backbone network for all experiments by standard mini-batch stochastic gradient descent (SGD) with momentum of $0.9$, weight decay of $2\times{10}^{-4}$. We train all the models on a single NVIDIA 1080Ti GPU for $200$ epochs with batch size of $128$. The initial learning rate is set to $0.1$ and the first five epochs is trained with the linear warm-up learning rate schedule~\cite{goyal2017accurate}. The learning rate is decayed at the $120^{\rm th}$ and $160^{\rm th}$ epoch by $0.01$ for our BBN.

{\noindent \textbf{Implementation details on iNaturalist. }}  
For fair comparisons, we utilize ResNet-50~\cite{he2016deep} as our backbone network in all experiments on iNaturalist 2017 and iNaturalist 2018. We follow the same training strategy in~\cite{goyal2017accurate} with batch size of $128$ on four GPUs of NVIDIA 1080Ti. We firstly resize the image by setting the shorter side to $256$ pixels and then take a $224 \times 224$ crop from it or its horizontal flip. During training, we decay the learning rate at the $60^{\rm th}$ and $80^{\rm th}$ epoch by $0.1$ for our BBN, respectively. 
% Note that we also train our method with $2\times$ scheduler and achieve much better performance. For fair comparisons, we also train the other state-of-the-art methods with $2\times$ scheduler.	

\subsection{Comparison methods}
In experiments, we compare our BBN model with three groups of methods:
%The methods compared with our proposed BBN consist of four types: the baseline methods, the two-stage methods, the state-of-the-art methods concerning long-tailed problems and the $2\times$ scheduler methods.

\begin{itemize}[itemsep=-0.2em, leftmargin=1em]
	\item {\noindent \textbf{Baseline methods.}} We employ plaining training with cross-entropy loss and focal loss~\cite{focalloss} as our baselines. Note that, we also conduct experiments with a series of mixup algorithms~\cite{mixup,manifoldmixup} for comparisons.
	\item {\noindent \textbf{Two-stage fine-tuning strategies.}}
	To prove the effectiveness of our cumulative learning strategy, we also compare with the two-stage fine-tuning strategies proposed in previous state-of-the-art~\cite{ldam}. We train networks with cross-entropy (CE) on imbalanced data in the first stage, and then conduct class re-balancing training in the second stage. ``CE-DRW'' and ``CE-DRS'' refer to the two-stage baselines using re-weighting and re-sampling at the second stage.
	\item {\noindent \textbf{State-of-the-art methods.}} For state-of-the-art methods, we compare with the recently proposed LDAM~\cite{ldam} and CB-Focal~\cite{cb-focal} which achieve good classification accuracy on these four aforementioned long-tailed datasets.
\end{itemize}

\begin{table}[t]
	\footnotesize
	\renewcommand\arraystretch{1.1}
	\caption{Top-1 error rates of ResNet-50 on large-scale long-tailed datasets iNaturalist 2018 and iNaturalist 2017. Our method outperforms the previous state-of-the-arts by a large margin, especially with {$2\times$} scheduler. ``*'' indicates original results in that paper.}
	\vspace{2mm}
	\centering
	\begin{tabular}{|c|c|c|}
		\hline
		Dataset        & iNaturalist 2018 & iNaturalist 2017           \\ \hline\hline
		CE              & 42.84  & 45.38            \\ 
		\hline
		CE-DRW~\cite{ldam}          & 36.27  &  40.48          \\ 
		CE-DRS~\cite{ldam}        & 36.44  &   40.12          \\ 
		\hline
		CB-Focal~\cite{cb-focal}        & 38.88  & 41.92            \\ 
		LDAM-DRW*~\cite{ldam}       & 32.00  & --            \\
		LDAM-DRW~\cite{ldam}        & 35.42  & 39.49            \\
		LDAM-DRW~\cite{ldam} ($2\times$)    & 33.88  & 38.19 \\
		\hline
		\hline
		Our BBN             & {33.71}  & {36.61}             \\
		Our BBN ($2\times$)         & \textbf{30.38}  &  \textbf{34.25}     \\ \hline
	\end{tabular}
	\label{tab:inat_results}
\end{table}

\subsection{Main results}

\subsubsection{Experimental results on long-tailed CIFAR}

We conduct extensive experiments on long-tailed CIFAR datasets with three different imbalanced ratios: $10$, $50$ and $100$. Table~\ref{tab:cifar_results} reports the error rates of various methods. We demonstrate that our BBN consistently achieves the best results across all the datasets, when comparing other comparison methods, including the two-stage fine-tuning strategies (\ie, CE-DRW/CE-DRS), the series of mixup algorithms (\ie, mixup, manifold mixup and manifold mixup with two samplers as the same as ours), and also previous state-of-the-arts (\ie, CB-Focal~\cite{cb-focal} and LDAM-DRW~\cite{ldam}).

Especially for long-tailed CIFAR-10 with imbalanced ratio $100$ (an extreme imbalance case), we get 20.18\% error rate which is 2.79\% lower than that of LDAM-DRW~\cite{ldam}. Additionally, it can be found from that table, the two-stage fine-tuning strategies (\ie, CE-DRW/CE-DRS) are effective, since they could obtain comparable or even better results comparing with state-of-the-art methods.

\subsubsection{Experimental results on iNaturalist}

Table~\ref{tab:inat_results} shows the results on two large-scale long-tailed datasets, \ie, iNaturalist 2018 and iNaturalist 2017. As shown in that table, the two-stage fine-tuning strategies (\ie, CE-DRW/CE-DRS) also perform well, which have consistent observations with those on long-tailed CIFAR. Compared with other methods, on iNaturalist, our BBN still outperform competing approaches and baselines. Besides, since iNaturalist is large-scale, we also conduct network training with the $2\times$ scheduler. Meanwhile, for fair comparisons, we further evaluate the previous state-of-the-art LDAM-DRW~\cite{ldam} with the $2\times$ training scheduler. It is obviously to see that, with $2\times$ scheduler, our BBN achieves significantly better results than BBN without $2\times$ scheduler. Additionally, compared with LDAM-DRW ($2\times$), we achieve $+3.50\%$ and $+3.94\%$ improvements on iNaturalist 2018 and iNaturalist 2017, respectively. In addition, even though we do not use the $2\times$ scheduler, our BBN can still get the best results. For a detail, we conducted the experiments based on LDAM~\cite{ldam} with the source codes provided by the authors, but failed to reproduce the results reported in that paper.

\subsection{Ablation studies}
\subsubsection{Different samplers for the re-balancing branch}

For better understanding our proposed BBN model, we conduct experiments on different samplers utilized in the re-balancing branch. We present the error rates of models trained with different samplers in Table~\ref{tab:sampler}. For clarity, the uniform sampler maintains the original long-tailed distribution. The balanced sampler assigns the same sampling possibility to all classes, and construct a mini-batch training data obeying a balanced label distribution. As shown in that table, the reversed sampler (our proposal) achieves considerably better performance than the uniform and balanced samplers, which indicates that the re-balancing branch of BBN should pay more attention to the tail classes by enjoying the reversed sampler.

\begin{table}[t]
	\footnotesize
	\caption{Ablation studies for different samplers for the re-balancing branch of BBN on long-tailed CIFAR-10-IR50.} 
	\vspace{2mm}
	\centering
	\begin{tabular}{|c|c|}
		\hline
		Sampler  &  Error rate \\ \hline\hline
		Uniform sampler &  21.31               \\ \hline 
		Balanced sampler &  21.06                 \\ \hline
		\hline
		Reversed sampler (Ours) &  \textbf{17.82}      \\ \hline
	\end{tabular}
	\label{tab:sampler}
\end{table}

\subsubsection{Different cumulative learning strategies}
\label{sec:alpha}

To facilitate the understanding of our proposed cumulative learning strategy, we explore several different strategies to generate the adaptive trade-off parameter $\alpha$ on CIFAR-10-IR50. Specifically, we test with both progress-relevant/irrelevant strategies, cf. Table~\ref{tab:adaptor}. For clarity, progress-relevant strategies adjust $\alpha$ with the number of training epochs, \eg, linear decay, cosine decay, \etc. Progress-irrelevant strategies include the equal weight or generate from a discrete distribution (\eg, the $\beta$-distribution).

\begin{table}[t]
\footnotesize
\renewcommand\arraystretch{1.5}
\caption{Ablation studies of different adaptor strategies of BBN on long-tailed CIFAR-10-IR50.} 
\vspace{2mm}
\centering
\begin{tabular}{|c|c|c|}
	\hline
	Adaptor  & $\alpha$   & Error rate \\ \hline\hline
	Equal weight & $0.5$     &    21.56              \\ \hline
	$\beta$-distribution & $Beta(0.2, 0.2)$     & 21.75                \\ \hline
	Parabolic increment & $\left(\frac{T}{T_{max}}\right)^{2}$  & 22.70      \\ \hline
	Linear decay & $1 - \frac{T}{T_{max}}$     &       18.55         \\ \hline
	Cosine decay & $\cos({\frac{T}{T_{max}}}\cdot\frac{\pi}{2})$     &    18.04             \\ \hline
	\hline
	Parabolic decay (Ours) & $1 - \left(\frac{T}{T_{max}}\right)^{2}$  & \textbf{17.82} \\ \hline
\end{tabular}
\label{tab:adaptor}
\end{table}

\iffalse
\begin{figure}[t]
	\centering
	\includegraphics[width=0.65\linewidth]{fig/alpha}
	%\vspace{-2mm}
	\caption{Different adaptor strategies for generating $\alpha$. The horizontal axis is current epoch ratio $\frac{T}{T_{max}}$ and the vertical axis indicates the value of $\alpha$.} 
	\label{fig:alpha}
\end{figure}
\fi

%It is a natural choice to set $\alpha$ as a constant 0.5 and our method can achieve $21.56\%$ error rate as shown in the first row of Table~\ref{tab:adaptor}. Then, we try the $\beta$-distribution, which is utilized in~\cite{mixup}, to sample values for $\alpha$ in the $\beta$-distribution, achieving the $21.75\%$ error rate. Above changing strategies are irrelevant to the progress of network training. 

%To prove our motivation that we should learn discriminative feature representations as the foundation for learning robust classifiers, we conduct ablation studies on different progress-relevant strategies, \ie, linear decay, cosine decay, parabolic decay and increment. Refer to Figure~\ref{fig:alpha} for visualization of different kinds for adaptors.

As shown in Table~\ref{tab:adaptor}, the decay strategies (\ie, linear decay, cosine decay and our parabolic decay) for generating $\alpha$ can yield better results than the other strategies (\ie, equal weight, $\beta$-distribution and parabolic increment). These observations prove our motivation that the conventional learning branch should be learned firstly and then the re-balancing branch. Among these strategies, the best way for generating $\alpha$ is the proposed parabolic decay approach. In addition, the parabolic increment, where re-balancing are attended before conventional learning, performs the worst, which validates our proposal from another perspective. More detailed discussions can be found in the supplementary materials.

\subsection{Validation experiments of our proposals}

\subsubsection{Evaluations of feature quality}

%To evaluate the representation quality of our proposed BBN, we extract the feature representations by the conventional learning branch and the re-balancing branch and train linear classifiers with CE on top of these feature representations on CIFAR-100-IR50.

It is proven in Section~\ref{sec:how class re-balancing work} that learning with vanilla CE on original data distribution can obtain good feature representations. 
In this subsection, we further explore the representation quality of our proposed BBN by following the empirical settings in Section~\ref{sec:how class re-balancing work}. Concretely, given a BBN model trained on CIFAR-100-IR50, firstly, we fix the parameters of representation learning of two branches. Then, we separately retrain the corresponding classifiers from scratch of two branches also on CIFAR-100-IR50. Finally, classification error rates are tested on these two branches independently.
%Ideally, owing to that the learning manners of classifiers are identical, more powerful representation ability of training features will lead to better classification performance.

As shown in Table~\ref{tab:representation quality of BBN}, the feature representations obtained by the conventional learning branch of BBN (``BBN-CB'') achieves comparable performance with CE, which indicates that our proposed BBN greatly preserves the representation capacity learned from the original long-tailed dataset. 
Note that, the re-balancing branch of BBN (``BBN-RB'') also gets better performance than RW/RS and it possibly owes to the parameters sharing design of our model.

\begin{table}[t]
	\footnotesize
	\caption{Feature quality evaluation for different learning manners.
	%linear classifiers trained on different kinds of feature extractors. The feature extractors and classifiers are all trained on CIFAR-100-IR50. BBN-CB and BBN-RB indicate the feature extractor of the conventional learning branch and the re-balancing branch respectively.
	}
	\vspace{2mm}
	\centering
	\begin{tabular}{|cc|}
		\hline
		\multicolumn{1}{|c|}{Representation learning manner} & Error rate \\ \hline\hline
		\multicolumn{1}{|c|}{CE}               & \textbf{58.62}      \\
		\multicolumn{1}{|c|}{RW}               & 63.17      \\
		\multicolumn{1}{|c|}{RS}               & 63.71      \\ \hline\hline
		\multicolumn{1}{|c|}{BBN-CB}           & 58.89      \\
		\multicolumn{1}{|c|}{BBN-RB}           & 61.09      \\ \hline
	\end{tabular}
	\label{tab:representation quality of BBN}
\end{table}

\subsubsection{Visualization of classifier weights}

Let denote $\bm{W} \in \mathbb{R}^{D\times C}$ as a set of classifiers $\{\bm{w}_1,\bm{w}_2,..., \bm{w}_C\}$ for all the $C$ classes, where $\bm{w}_i \in \mathbb{R}^{D}$ indicates the weight vector for class $i$. Previous work~\cite{guo2017one} has shown that the value of $\ell_2$-norm ${\left\lbrace\|\bm{w}_i\|_2\right\rbrace }_{i=1}^{C}$ for different classes can demonstrate the preference of a classifier, \ie, the classifier $\bm{w}_i$ with the largest $\ell_2$-norm tends to judge one example as belonging to its class $i$. Following~\cite{guo2017one}, we visualize the $\ell_2$-norm of these classifiers.

\begin{figure}[t]
	\centering
	\includegraphics[width=0.695\linewidth]{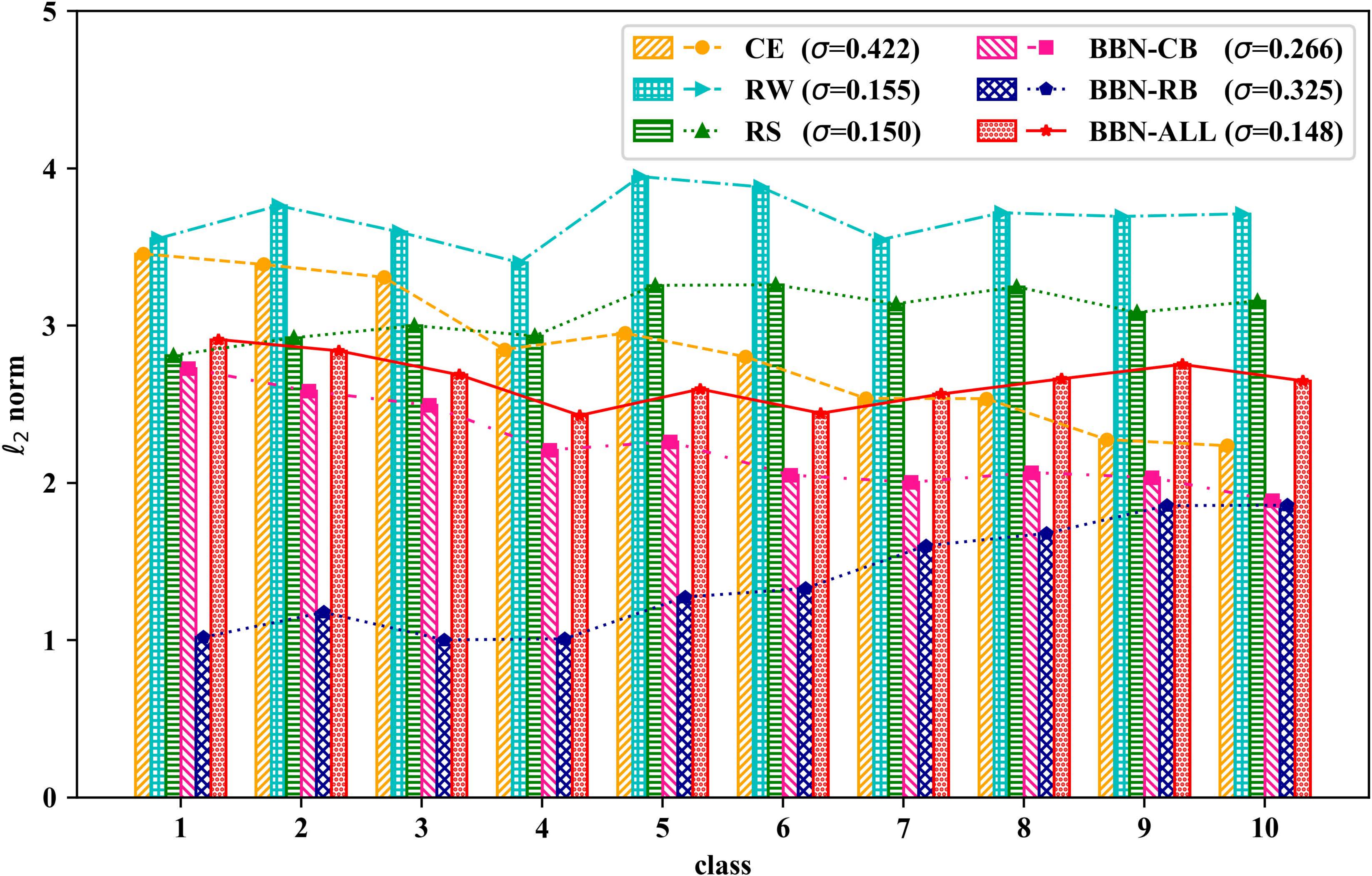}
	%\vspace{1mm}
	\caption{$\ell_2$-norm of classifier weights for different learning manners. Specifically, ``BBN-ALL'' indicates the $\ell_2$-norm of the combination of $\bm{W}_c$ and $\bm{W}_r$ in our model. $\sigma$ in the legend is the standard deviation of $\ell_2$-norm for ten classes.}
	%\vspace{-1em}
	\label{fig:weight vis}
\end{figure}

%Specifically, the $\ell_2$-norms learned with the uniform sampler (BBN-CB) is still in a long-tail distribution while $\ell_2$-norms learned with the reversed sampler (BBN-RB) are in a reversed long-tailed distribution.

As shown in Figure~\ref{fig:weight vis}, we visualize the $\ell_2$-norm of ten classes trained on CIFAR-10-IR50. For our BBN, we visualize the classifier weights $\bm{W}_{c}$ of the conventional learning branch (``BBN-CB'') and the classifier weights $\bm{W}_{r}$ of the re-balancing branch (``BBN-RB''), as well as their combined classifier weights (``BBN-ALL''). Additionally, the visualization results on classifiers trained with these learning manners in Section~\ref{sec:how class re-balancing work}, \ie, CE, RW and RS, are also provided.

Obviously, the $\ell_2$-norm of ten classes' classifiers for our proposed model (\ie, ``BBN-ALL'') are basically equal, and their standard deviation $\sigma=0.148$ is the smallest one. For the classifiers trained by other learning manners, the distribution of the $\ell_2$-norm of CE is consistent with the long-tailed distribution. The $\ell_2$-norm distribution of RW/RS looks a bit flat, but their standard deviations are larger than ours. It gives an explanation why our BBN can outperform these methods. Additionally, by separately analyzing our model, its conventional learning branch (``BBN-CB'') has a similar $\ell_2$-norm distribution with CE's, which justifies its duty is focusing on universal feature learning. The $\ell_2$-norm distribution of the re-balancing branch (``BBN-RB'') has a reversed distribution w.r.t. original long-tailed distributions, which reveals it is able to model the tail.

\section{Conclusions}

In this paper, for studying long-tailed problems, we explored how class re-balancing strategies influenced representation learning and classifier learning of deep networks, and revealed that they can promote classifier learning significantly but also damage representation learning to some extent. Motivated by this, we proposed a Bilateral-Branch Network (BBN) with a specific cumulative learning strategy to take care of both representation learning and classifier learning for exhaustively improving the recognition performance of long-tailed tasks. By conducting extensive experiments, we proved that our BBN could achieve the best results on long-tailed benchmarks, including the large-scale iNaturalist. In the future, we attempt to tackle the long-tailed detection problems with our BBN model.

\clearpage

\onecolumn
\appendix
\section*{SUPPLEMENTARY MATERIALS}
In the supplementary materials, we provide more experimental results and analyses of our proposed BBN model, including:

\begin{enumerate}
	\item[A.] Additional experiments of different manners for representation and classifier learning (cf. Section 3 and Figure 2 of the paper) on large-scale datasets iNaturalist 2017 and iNaturalist 2018;
	\item[B.] Affects of re-balancing strategies on the compactness of learned features;
	\item[C.] Comparisons between the BBN model and ensemble methods;
	\item[D.] Coordinate graph of different adaptor strategies for generating $\alpha$;
	\item[E.] Learning algorithm of our proposed BBN model.
\end{enumerate}

\newpage

\section{Additional experiments of different manners for representation and classifier learning (cf. Section 3 and Figure 2 of the paper) on large-scale datasets iNaturalist 2017 and iNaturalist 2018}
%1.ce，rw，rs在大数据集上的9宫格，17训feature，然后18和17分别训classifier
In this section, following Section~3 of our paper, we conduct experiments on large-scale datasets, \ie, iNaturalist 2017~\cite{van2018inaturalist} and iNaturalist 2018, to further justify our conjecture (\ie, the working mechanism of these class re-balancing strategies is to promote classifier learning significantly but might damage the universal representative ability of the learned deep features due to distorting original distributions.) Specifically, the representation learning stages are conducted on iNaturalist 2017. Then, to also evaluate the generalization ability for learned representations, classifier learning stages are performed on not only iNaturalist 2017 but also iNaturalist 2018.

%Top-1 error rates of different manners for representation learning and classifier learning are reported on large-scale long-tailed benchmark datasets iNaturalist 2017~\cite{van2018inaturalist} and iNaturalist 2018. 

As shown in Figure~\ref{fig:representation and classifier learning inat} of the supplementary materials, we can also have the observations from two perspectives on these large-scale long-tailed datasets:

%the representations trained with CE get lower error rates than those with RW/RS, while the error rates of classifiers trained with RW/RS are lower than CE. The conclusion is consistent with we argued in the paper that the way these strategies work is to promote classifier learning significantly but will damage the universal representative ability of the learned deep features to some extent.

\begin{itemize}%[itemsep=-0.2em, leftmargin=1em]
\item \textbf{Classifiers:} When we apply the same representation learning manner (comparing error rates of three blocks in the vertical direction), it can be reasonably found that RW/RS always achieve lower classification error rates than CE, which owes to their re-balancing operations adjusting the classifier weights updating to match test distributions.
\item \textbf{Representations:} When applying the same classifier learning manner (comparing error rates of three blocks in the horizontal direction), it is a bit of surprise to see that error rates of CE blocks are consistently lower than error rates of RW/RS blocks. The findings indicate that training with CE achieves better classification results since it obtains better features. The worse results of RW/RS reveal that they lead to inferior discriminative ability of the learned deep features.
\end{itemize}

These observations are consistent with those on long-tailed CIFAR datasets, which can further demonstrate our discovery of Section~3 in the paper.

\begin{figure*}[h]
	%\vspace{-4mm}
	\centering
	\begin{minipage}[t]{0.48\textwidth}
		\centering
		\includegraphics[width=6cm]{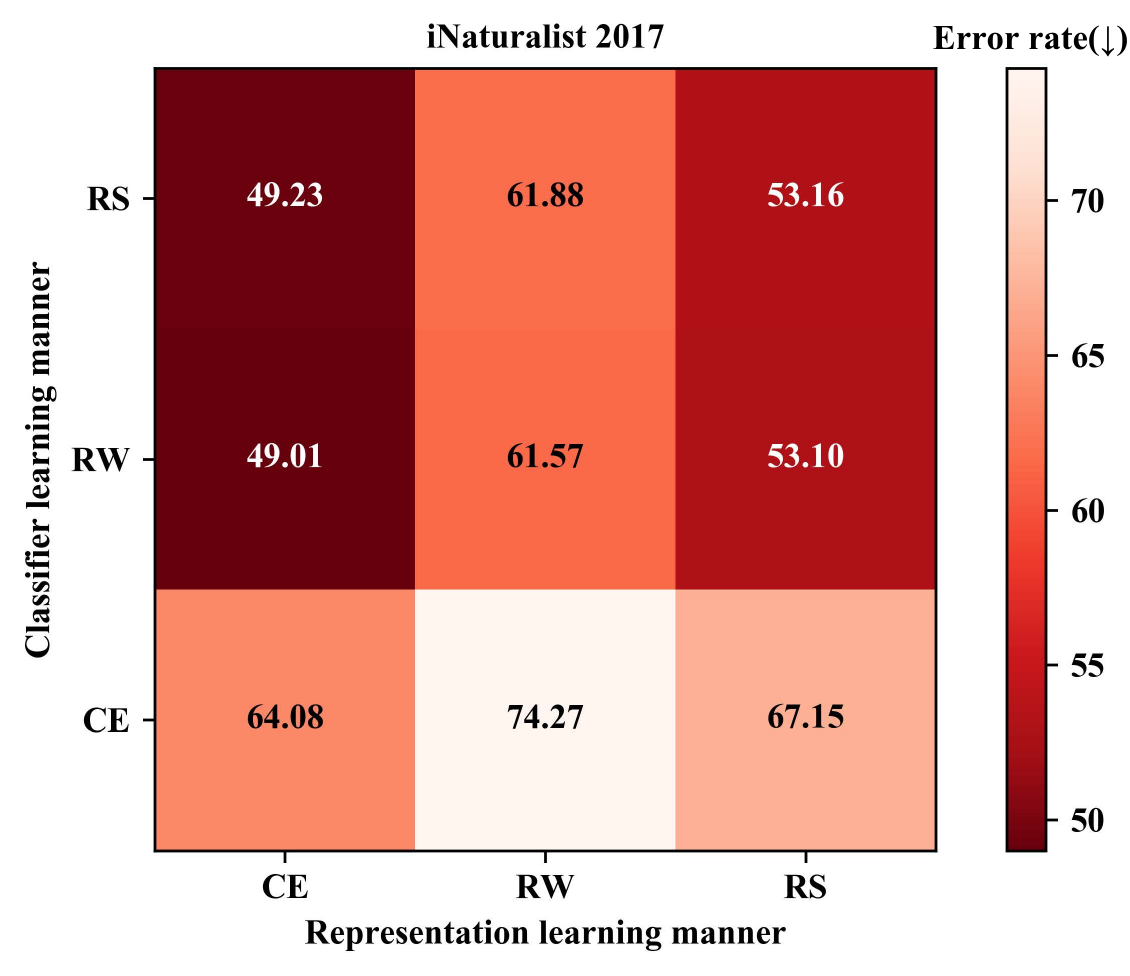}
	\end{minipage}
	\begin{minipage}[t]{0.48\textwidth}
		\centering
		\includegraphics[width=6cm]{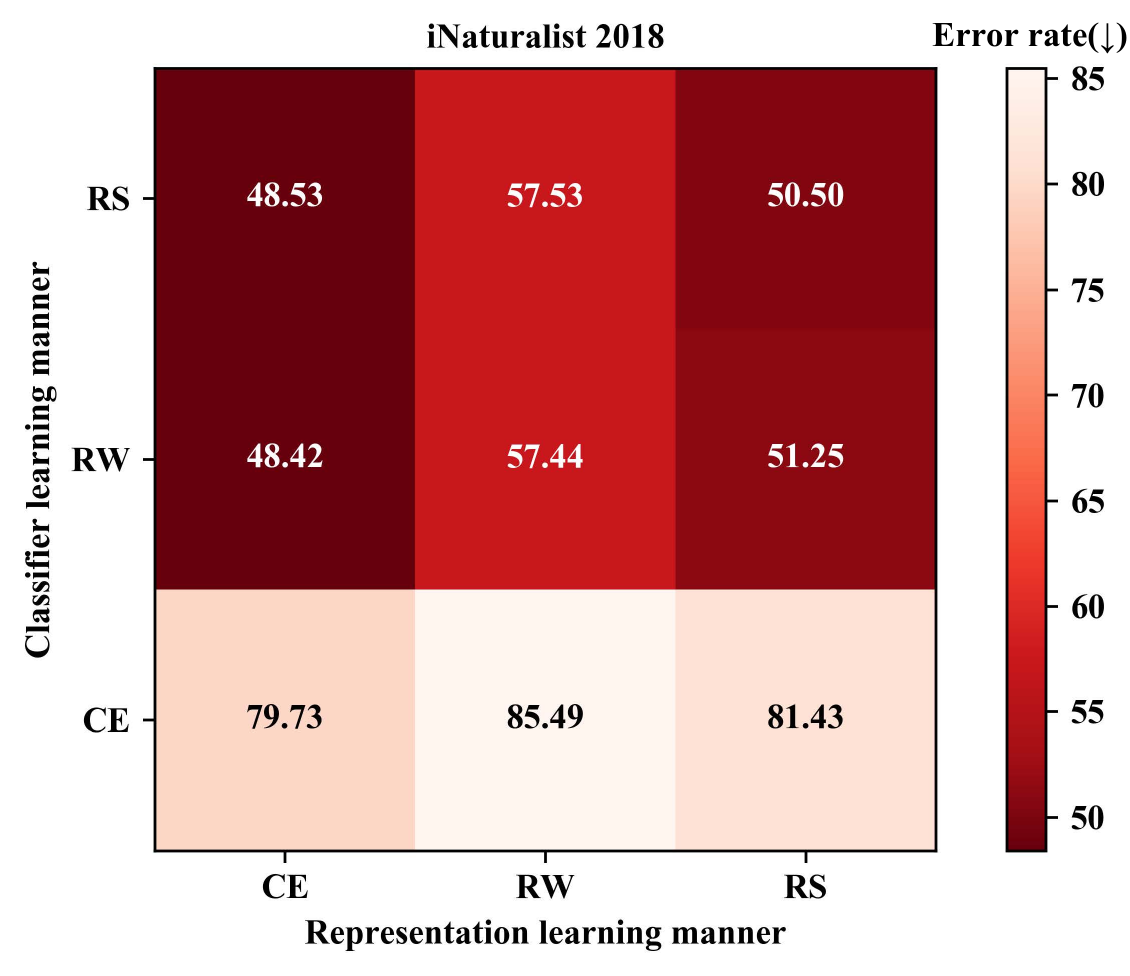}
	\end{minipage}
	\caption{Top-1 error rates of different manners for representation learning and classifier learning on two large-scale long-tailed datasets iNaturalist 2017 and iNaturalist 2018. “CE” (Cross-Entropy), “RW” (Re-Weighting) and “RS” (Re-Sampling) are the conducted learning manners.}
	\label{fig:representation and classifier learning inat}
\end{figure*}

%Specifically, ``CE'' (Cross-Entropy), ``RW'' (Re-Weighting) and ``RS'' (Re-Sampling) are the conducted learning manners in experiments. As observed, the representations trained with CE get lower error rates than those with RW/RS, while the error rates of classifiers trained with RW/RS are lower than CE.

\section{Affects of re-balancing strategies on the compactness of learned features}
%2.每个类与类中心的距离，直方图
%Compact feature representations are beneficial to the classification performance. 

To further prove our conjecture that re-balancing strategies could damage the universal representations, we measure the compactness of intra-class representations on CIFAR-10-IR50~\cite{cifar} for verification.

Concretely, for each class, we firstly calculate a centroid vector by averaging representations of this class. Then, $\ell_2$ distances between these representations and their centroid are computed and then averaged as a measurement for the compactness of intra-class representations. If the averaged distance of a class is small, it implies that representations of this class gather closely in the feature space. We normalize the $\ell_2$-norm of representations to $1$ in the training stage for avoiding the impact of feature scales. We report results based on representations learned with Cross-Entropy (CE), Re-Weighting (RW) and Re-Sampling (RS), respectively.

As shown in Figure~\ref{fig:distance} of the supplementary materials, the averaged distances of re-balancing strategies are obviously larger than conventional training, especially for the head classes. That is to say, the compactness of learned features of re-balancing strategies are significantly worse than conventional training. These observations can further validate the statements in Figure~1 of the paper (\ie, for re-balancing strategies, ``the intra-class distribution of each class becomes more separable") and also the discovery of Section~3 in the paper (\ie, re-balancing strategies ``might damage the universal representative ability of the learned deep features to some extent'').
%It, to some extent, illustrates that the compact property of representations on tail classes learned by cross-entropy is damaged.

\begin{figure*}[t!]
	\centering
	\includegraphics[width=0.7\linewidth]{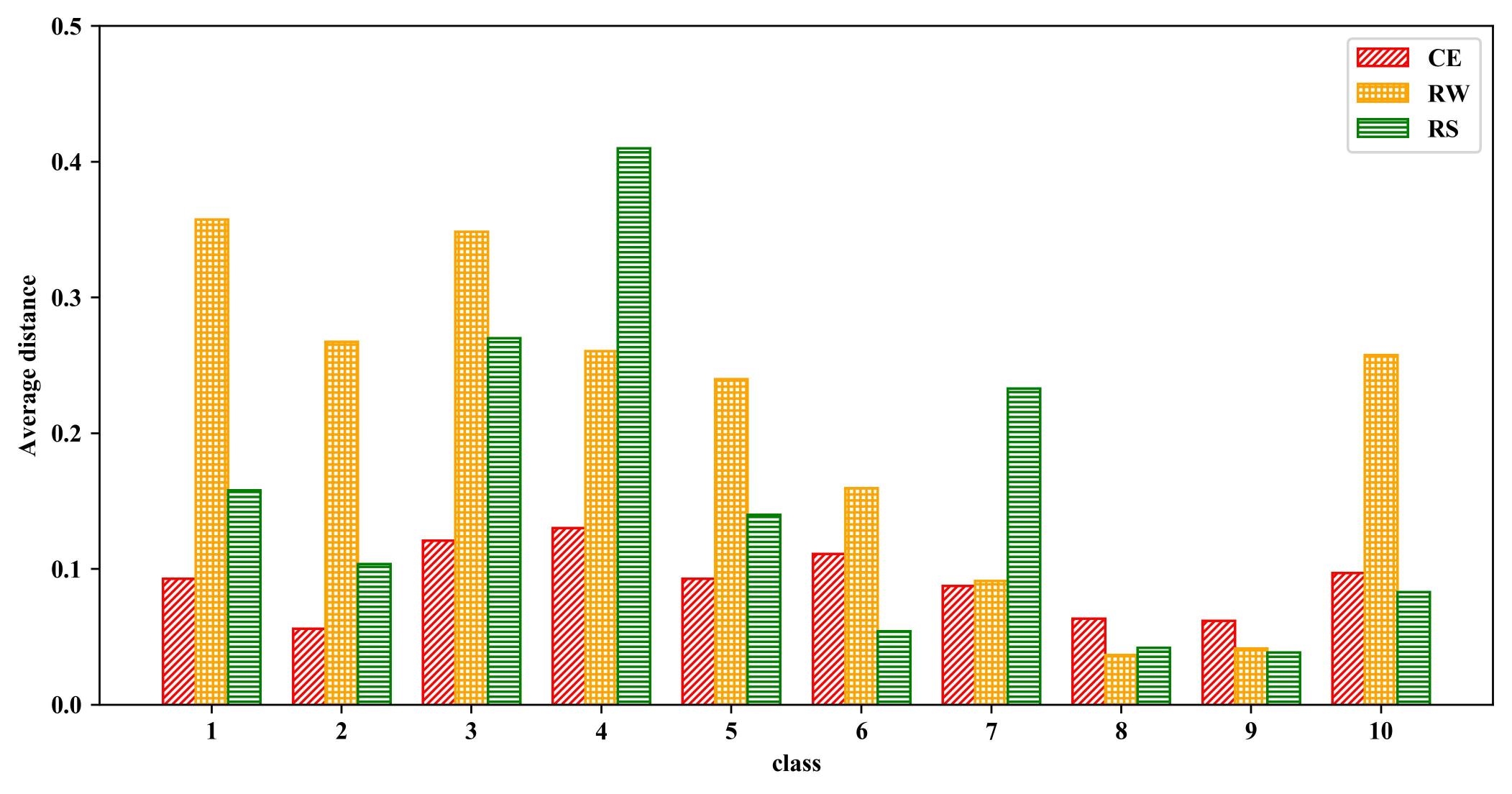}
	%\vspace{-2mm}
	\caption{Histogram of the measurement for the compactness of intra-class representations on the CIFAR-10-IR50 dataset. Especially for head classes, representations trained with CE gather more closely than those trained with RW/RS, since the representations of each class are closer to their centroid. The vertical axis is the averaged distance between learned features of each class and their corresponding centroid (The smaller, the better).}
	\label{fig:distance}
\end{figure*}

\section{Comparisons between the BBN model and ensemble methods}
%3.ensemble结果对比，4个数据集的（cifar10，cifar100，inat17，inat18）
%是不是可以直接说，为了验证我们在参数量少很多的情况下，结果反而大于ensemble
In the following, we compare our BBN model with ensemble methods to prove the effectiveness of our proposed model. Results on CIFAR-10-IR50~\cite{cifar}, CIFAR-100-IR50~\cite{cifar}, iNaturalist 2017~\cite{van2018inaturalist} and iNaturalist 2018 are provided in Table~\ref{tab:ensemble} for comprehensiveness. 

As known, ensemble techniques are frequently utilized to boost performances of machine learning tasks. We train three classification models with uniform data sampler, balanced data sampler and reversed data sampler, respectively. For mimicking our bilateral-branch network design and considering fair comparisons, we provide classification error rates of (1) an ensemble of models learned with a uniform sampler and a balanced sampler, as well as (2) another ensemble of models learned with a uniform sampler and a reversed sampler. 

As shown in Table~\ref{tab:ensemble} of the supplementary materials, our BBN model achieves consistently lower error rates than ensemble models on all datasets. Additionally, compared to ensemble models, our proposed BBN model can yield better performance with limited increase of network parameters thanks to its sharing weights design (cf. Sec.~4.2 of the paper).

\begin{table*}[t!]
	\centering
	\small
	\caption{Top-1 error rates of our proposed BBN model and ensemble methods.}
	\vspace{2mm}
	\label{tab:ensemble}
	\begin{tabular}{|c|c|c|c|c|}
		\hline
		Methods                                                               & CIFAR-10-IR50 & CIFAR-100-IR50 & iNaturalist 2017 & iNaturalist 2018 \\ \hline \hline
		\begin{tabular}[c]{@{}c@{}}Uniform sampler + Balanced sampler\end{tabular} & 19.41         & 55.10          & 39.53           & 36.20          \\
		\begin{tabular}[c]{@{}c@{}}Uniform sampler + Reversed sampler\end{tabular} & 19.38         & 54.93          & 40.02           & 36.66           \\ \hline \hline
		BBN (Ours)                                                                  & \textbf{17.82}         & \textbf{52.98}          & \textbf{36.61}           & \textbf{33.74}           \\ \hline
	\end{tabular}
\end{table*}

\section{Coordinate graph of different adaptor strategies for generating $\alpha$}
%4.alpha变化图，以前就有
As shown in Figure~\ref{fig:alpha} of the supplementary materials, we provide a coordinate graph to present how $\alpha$ varies with the progress of network training. The adaptor strategies shown in the figure are the same as those in Table 4 of the paper except for the $\beta$-distribution for its randomness.

Furthermore, as discussed in Sec.~5.5.2 of the paper, these decay strategies yield better results than the other non-decay strategies. When $\alpha$ decreasing, the learning focus of our BBN gradually changes from representation learning to classifier learning, which fits our motivation stated in Sec.~4.3 of the paper. Among these decay strategies, our proposed parabolic decay is the best. Specifically, we can intuitively regard $\alpha>0.5$ as the learning focus emphasizing representation learning, as well as $\alpha \leq 0.5$ as the learning focus emphasizing classifier learning. As shown in Figure~\ref{fig:alpha} of the supplementary materials, compared with other decay strategies, our parabolic decay \textit{with the maximum degree} prolongs the epochs of the learning focus upon representation learning. As analyzed by theoretical understanding of learning dynamics in networks~\cite{theory}, network convergence speed is highly correlated with the number of layers. That is to say, the representation learning part (former layers) of networks requires more epochs to sufficiently converge, while the classifier learning part (later layers) requires relatively less epochs until sufficient convergence. In fact, our parabolic decay ensures that BBN could have enough epochs to fully update the representation learning part, \ie, learning better universal features, which is the crucial foundation for learning robust classifiers. That is why our parabolic decay is the best.% We will add these discussions in the final version.

\begin{figure*}[t!]
	\centering
	\includegraphics[width=0.4\linewidth]{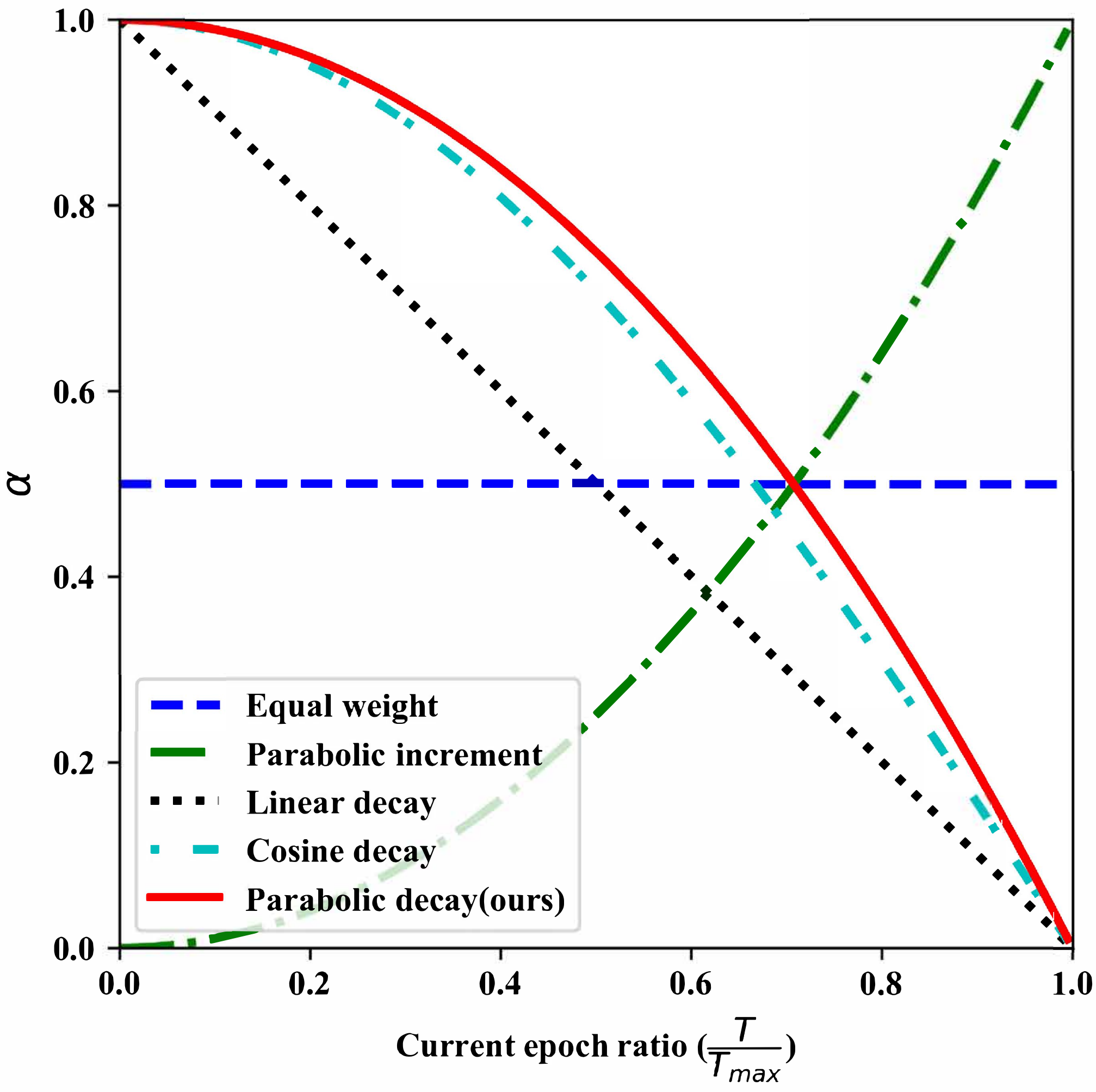}
	%\vspace{-2mm}
	\caption{Different kinds of adaptor strategies for generating $\alpha$. The horizontal axis indicates current epoch ratio $\frac{T}{T_{max}}$ and the vertical axis denotes the value of $\alpha$. (Best viewed in color)}
	\label{fig:alpha}
\end{figure*}

\section{Learning algorithm of our proposed BBN model}
%5.alg表，以前就有
In the following, we provide the detailed learning algorithm of our proposed BBN. In Algorithm~\ref{alg:bbn} of the supplementary materials, for each training epoch $T$, we firstly assign a value to $\alpha$ by the adaptor proposed in Eq.~(5) of the paper. Then, we sample training samples by the uniform sampler and reversed sampler, respectively. After feeding samples into our network, we can obtain two independent feature vectors $\bm{f}_c$ and $\bm{f}_r$. Then, we calculate the output logits $\bm{z}$ and the prediction possibility $\hat{\bm{p}}$ according to Eq.~(1) and Eq.~(2) of the paper. Finally, the classification loss function is calculated based on Eq.~(3) of the paper and we update model parameters by optimizing this loss function.

\begin{algorithm}[h!]
 \small
 \caption{Learning algorithm of our proposed BBN}
 {\textbf{Require :} Training Dataset $\mathcal{D}$ = ${\left\lbrace(\mathbf{x}_i,y_i)\right\rbrace }_{i=1}^{n}$; $\texttt{UniformSampler}(\cdot)$ denotes obtaining a sample from  $\mathcal{D}$ selected by a uniform sampler; $\texttt{ReversedSampler}(\cdot)$ denotes obtaining a sample by a reversed sampler; $\mathcal{F}_{cnn}(\cdot;\cdot)$ denotes extracting the feature representation from a CNN; $\theta_c$ and  $\theta_r$ denote the model parameters of the conventional learning and re-balancing branch; $\bm{W}_c$ and $\bm{W}_r$ present the classifiers' weights (\ie, last fully connected layers) of the conventional learning and re-balancing branch.}
 \begin{algorithmic}[1]
  \FOR{$T$ = 1 to $T_{max}$} % For 语句，需要和EndFor对应
  \STATE $\alpha \leftarrow 1 - \left(\frac{T}{T_{max}}\right)^{2}$
  \STATE $(\mathbf{x}_c, y_c) \leftarrow \texttt{UniformSampler}(\mathcal{D})$ 
  \STATE $(\mathbf{x}_r, y_r) \leftarrow \texttt{ReversedSampler}(\mathcal{D})$
  \STATE $\bm{f}_c \leftarrow \mathcal{F}_{cnn}(\mathbf{x}_c; \theta_c)$
  \STATE $\bm{f}_r \leftarrow \mathcal{F}_{cnn}(\mathbf{x}_r; \theta_r)$
  \STATE $\mathbf{z} \leftarrow \alpha  \bm{W}^{\top}_c  \bm{f}_c + (1-\alpha)  \bm{W}^{\top}_r  \bm{f}_r$
  \STATE $\bm{\hat{p}} \leftarrow  \texttt{Softmax}(\bm{z})$
  \STATE $\mathcal{L} \leftarrow \alpha  E({\bm{\hat{p}}, y_c}) + (1-\alpha)  E({\bm{\hat{p}}, y_r})$
  \STATE Update model parameters by minimizing $\mathcal{L}$
  \ENDFOR
 \end{algorithmic}
 \label{alg:bbn}
\end{algorithm}

\clearpage
\twocolumn
{\small
\bibliographystyle{ieee_fullname}
\bibliography{egbib}
}

\end{document}